# Deep Learning with Lung Segmentation and Bone Shadow Exclusion Techniques for Chest X-Ray Analysis of Lung Cancer

Yu.Gordienko[1*1], Peng Gang[2], Jiang Hui[2], Wei Zeng[2], Yu.Kochura[1], O.Alienin[1], O. Rokovyi[1], and S. Stirenko[1]

[1]National Technical University of Ukraine "Igor Sikorsky Kyiv Polytechnic Institute", Kyiv, Ukraine
`{yuri.gordienko,sergii.stirenko,iuriy.kochura}@gmail.com`
`http://comsys.kpi.ua`
[2]Huizhou University, Huizhou City, China
`peng@hzu.edu.cn`
`http://www.hzu.edu.cn`

**Abstract.** The recent progress of computing, machine learning, and especially deep learning, for image recognition brings a meaningful effect for automatic detection of various diseases from chest X-ray images (CXRs). Here efficiency of lung segmentation and bone shadow exclusion techniques is demonstrated for analysis of 2D CXRs by deep learning approach to help radiologists identify suspicious lesions and nodules in lung cancer patients. Training and validation was performed on the original JSRT dataset (dataset #01), BSE-JSRT dataset, i.e. the same JSRT dataset, but without clavicle and rib shadows (dataset #02), original JSRT dataset after segmentation (dataset #03), and BSE-JSRT dataset after segmentation (dataset #04). The results demonstrate the high efficiency and usefulness of the considered pre-processing techniques in the simplified configuration even. The pre-processed dataset without bones (dataset #02) demonstrates the much better accuracy and loss results in comparison to the other pre-processed datasets after lung segmentation (datasets #02 and #03).

## 1 Introduction

Lung cancer is a significant burden in the world and especially in China, where as more than half of adult men in that country are and heavy air pollution aggravate the consequences of the disease. Lung cancer in the world and China has a very high disease incidence and the current solution consists in early screening which often leads to good outcomes at a relatively low cost. Screening, especially by computed tomography, has been recognized worldwide as an approach to reduce lung cancer mortality [1]. But its relatively low outspread is aggravated by the high cost and limited availability for the most parts of the world. Chest X-ray (CXR) imaging is currently the most popular and the most available diagnostic tool for health monitoring and diagnosing many lung diseases, including pneumonia, tuberculosis, cancer, etc. However, detecting marks of these diseases from CXRs is a very complicated procedure, which takes involvement of the expert radiologists. Application of CXRs is postponed by long manual analysis and detection of lung cancer and it is limited also by shortage of experts. For example, in China the annual number of the diagnosed lung cancer cases is huge (>600 thousands), but the number of certified radiologists is low (<80 thousands) for the nation-wide screening of >1.4 billion of citizens.

---
[1] Corresponding author.

Meanwhile, the recent disruptive progress of computing, especially computing on the general purpose graphic processing units (GPU) [2,3], machine learning, and especially deep learning [4], for image recognition brings a meaningful effect. For example, recently, CheXNet model was announced that can automatically detect pneumonia from chest X-rays at a level exceeding practicing radiologists [5]. That is why any automated assistance tools and related machine learning techniques are of great importance for the faster and better identification, classification and segmentation of suspicious regions (like lesions, nodules, etc.) for the subsequent diagnostic.

The main aim of this paper is to demonstrate efficiency of lung segmentation and bone shadow exclusion techniques for analysis of 2D CXRs by deep learning approach to help radiologists identify suspicious regions in lung cancer patients.

## 2    Problem and Related Work

Several screening approaches are used now to detect suspicious lung cancer lesions. Computed tomography (CT) is especially sensitive to hard-to-detect nodules and enhances radiologists' diagnosis accuracy. X-ray is often thought as an obsolete medical imaging method, but usage of digital technologies and machine learning now revives the significance of X-ray in medical imaging diagnosis. For example, they allow to detect more different kinds of cardiothoracic lesions and especially sensitive to lung nodules on X-ray scans. The success is strengthened by the fast progress in machine learning and GPU computing research for medical data processing. Recently, several promising results were obtained in the field of lung diseases diagnostic by machine learning and, especially, by deep learning approaches. As a result, the current GPU-based platform can process hundreds of high-resolution medical images per second. Availability of the open datasets with CT and CXR images allow data scientists to train, verify, and tune their new algorithms. The release of the image database with and without lung cancer nodules proposed by Japanese Society of Radiological Technology (JSRT) open this way for many research groups around the world [6]. This initiative was supported by other medical and research institutions. Lung Image Database Consortium (LIDC) proposes the image database containing data captured by various modalities (computed tomography — CT, digital radiography — DX, computed radiography — CR) for >1000 patients, and contain >244 000 images with in-plane resolution of the 512×512-pixel sections ranged from 0.542-0.750 mm [7]. The U.S. National Library of Medicine has made two open datasets of CXRs to foster research in computer-aided diagnosis of pulmonary diseases with a special focus on pulmonary tuberculosis [8,9]. Montgomery County

(MC) dataset has been collected in collaboration with the Department of Health and Human Services, Montgomery County in Maryland, USA. Shenzhen Hospital dataset (SH) was acquired from Shenzhen No. 3 People's Hospital in Shenzhen, China. Both datasets contain normal and abnormal chest X-rays with manifestations of tuberculosis and include associated radiologist readings. Now ChestX-ray14 is the largest publicly available chest X-ray dataset, containing over 100 000 frontal-view X-ray images with 14 different lung diseases. [10]. The aforementioned CheXNet model, which is a 121-layer convolutional neural network, was trained on ChestX-ray14 dataset [5]. To increase the accuracy of predictions researchers try to exclude the regions which are not pertinent to lungs or other regions of interest. Such task includes segmenting the left and right lung fields in standard CXRs. Several segmentation approaches were proposed recently, which are based on active shape models, active appearance models, and a multi-resolution pixel classification method. The methods have been tested on JSRT database, in which all objects have been manually segmented by two human observers [11]. Although many new segmentation methods have been proposed for CXR and MRI images [12] in medical image applications, the lung field segmentation remains a challenge.

Additional promising option for improvement of prediction is related with exclusion of some body parts that shadow the lung, for example, ribs and clavicles. As a service to the medical imaging community, Chest Diagnostic System Research Group (Budapest, Hungary) provided the bone shadow eliminated (BSE) version of JSRT dataset (BSE-JSRT) [13]. In other study, the two-step algorithm for eliminating rib shadows in CXRs was based on delineation of the ribs using a novel hybrid self-template approach and then suppression of these delineated ribs using an unsupervised regression model that takes into account the change in proximal thickness of bone in the vertical axis [14]. Below the further details are given for the datasets used in this work.

## 3 Data and Methods

### 3.1. Bone Elimination

Below the segmentation techniques were applied to datasets JSRT and BSE-JSRT datasets. JSRT image dataset contains 247 images (Fig. 1a): 154 cases with lung nodules and 93 cases without lung nodules [6]. BSE-JSRT dataset contains 247 images (Fig. 1b) of the JSRT dataset, but without clavicle and rib shadows removed by the special algorithms [13]. The one of the aims of the work was to check the difference between application of the deep learning approach to the original JSRT dataset (below it is mentioned as dataset #01) and BSE-JSRT dataset, i.e. the same JSRT dataset, but without clavicle and rib shadows (dataset #02).

### 3.2. Lung Segmentation

To perform lung segmentation, i.e. to cut the left and right lung fields from the heart and other parts in standard CXRs, the UNet-based convolutional neural network

(CNN) was applied. The UNet has CNN architecture for fast and precise segmentation of images, which demonstrated high accuracy on several challenges dedicated to segmentation of neuronal structures in electron microscopic stacks, detection of caries in bitewing radiography, and cell tracking from transmitted light microscopy [15]. Recently, this approach was successfully used for segmenting lungs on CXRs from MC and JSRT datasets with usage of manually prepared masks [16]. The additional aim of the work was to check the effect of lung segmentation for application of the deep learning approach to the original JSRT dataset after segmentation (below it is mentioned as dataset #03) and the same BSE-JSRT dataset after segmentation (below it is mentioned as dataset #04).

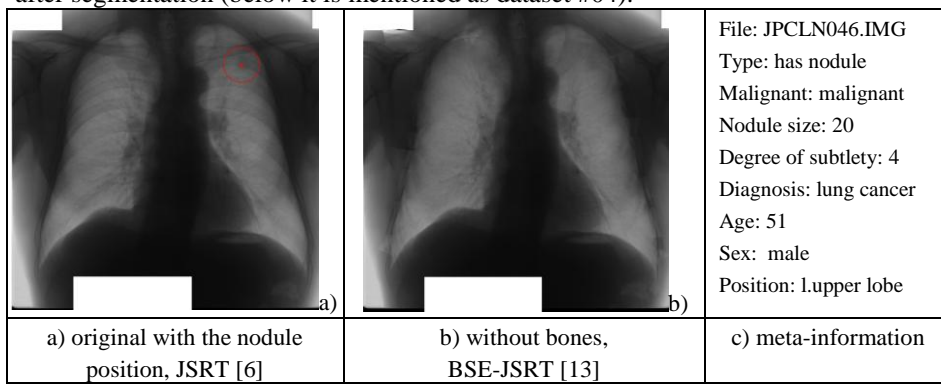

| a) original with the nodule position, JSRT [6] | b) without bones, BSE-JSRT [13] | c) meta-information |
|---|---|---|

**Fig. 1.** Example of the original image (2048×2048 pixels) with the cancer nodule from JSRT dataset [6] (a), the correspondent image without bones from BSE-JSRT dataset [13] (b), and the related meta-information (c). The nodule location and region are denoted by the point and circle respectively (a).

### 3.3. Model, Running Time and Speedup Analysis

The selection of training model was motivated by the reasons of simplicity and short running time to investigate the effect of segmentation and bone elimination only in the simplified configuration without the emphasize on the highest possible accuracy and lowest loss (which will be investigated in the further works). The simple and standard CNN model with only 7 convolution 2D layers was used for training on the original, segmented, and bone eliminated datasets which were mentioned above. The running time (Fig. 2) and speedup analysis (Fig. 3) was performed after numerous tests for images of various sizes and batch sizes in "single-CPU" (1 core of Intel i7), "multi-CPU" (8 cores of Intel i7), and "GPU" (graphic processing unit — NVIDIA Tesla K40c) modes. The obtained results demonstrated the maximal speedup of 3.0 times in multi-CPU mode and up to 9.5 times in GPU mode for the biggest image sizes (1024×1024) for the batch size of 8 images. These results allowed to estimate the realistic scenarios for the feasibility tests of segmentation and bone elimination techniques. Finally, the images of sizes 256×256 were selected for the previous training and reported here, and the training of the more complicated models on the bigger images are under work now and will be reported elsewhere [17].

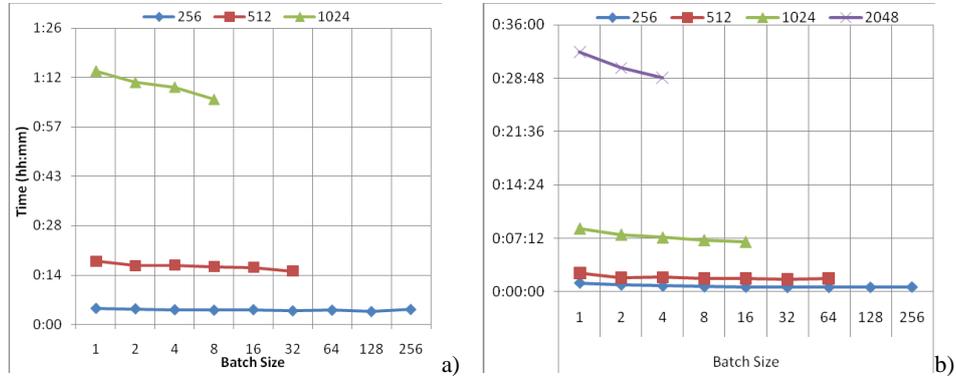

**Fig. 2.** Running time in CPU (a) and GPU (b) regimes for various batch and images sizes.

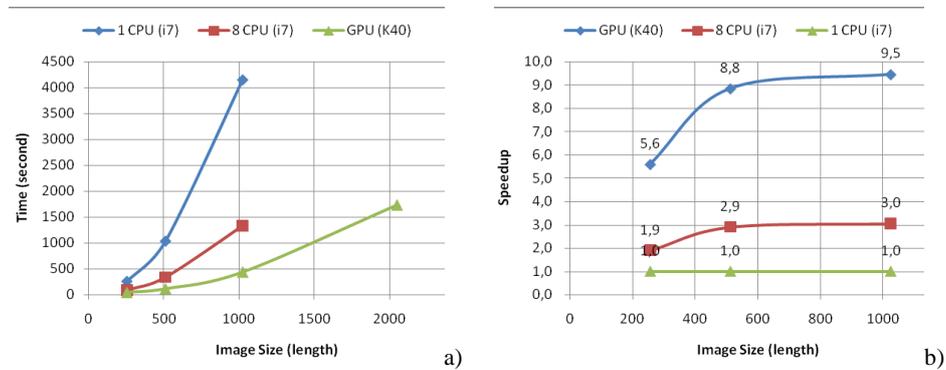

**Fig. 3.** Speedup of multi-CPU and GPU regimes in comparison to single-CPU mode

## 4    Results

In this section, the results are presented as to effect of bone elimination and lung segmentation on training with regard to: the original JSRT dataset (below it is mentioned as dataset #01), BSE-JSRT dataset, i.e. the same JSRT dataset, but without clavicle and rib shadows (dataset #02), original JSRT dataset after segmentation (dataset #03), and the same BSE-JSRT dataset after segmentation (dataset #04).

### 3.4.  Segmentation

This segmentation stage was applied to the original images from JSRT dataset (dataset #01, Fig. 4a) to obtain their segmented versions (dataset #03, Fig. 4c) and consisted in the following stages:
- training the UNet-based CNN for lung segmentation (search of lungs borders) on MC dataset with manually prepared masks (lung borders),
- predicting the lung borders in the shape of black-and-white lung masks (Fig. 4b) by means of the trained UNet-based CNN for each of original images from JSRT dataset (Fig. 4a),
- cutting the regions of interest (right and left lungs) (Fig. 4c) from their original images (Fig. 4a).

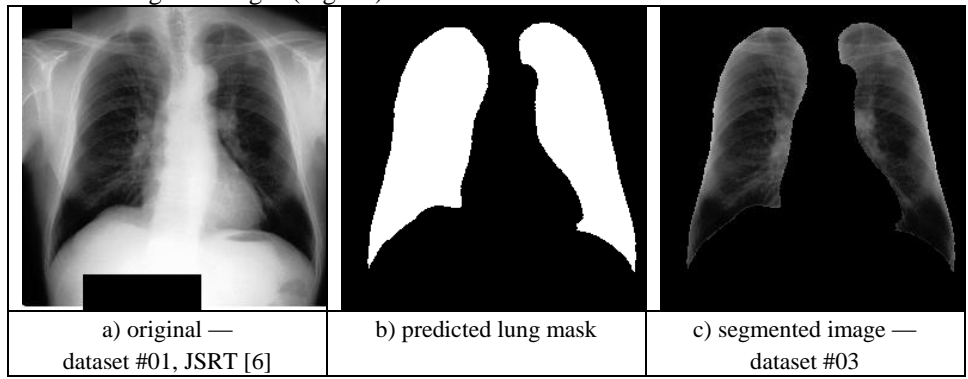

| a) original — dataset #01, JSRT [6] | b) predicted lung mask | c) segmented image — dataset #03 |

**Fig. 4.** Example of the original image (inversed version of Fig.1a) (a), the correspondent lung mask predicted by machine learning approach (b), and its segmented version (c).

The similar procedure was applied to the images without bones from BSE-JSRT dataset (dataset #02, Fig. 5a) to obtain their segmented versions without bones (dataset #04, Fig. 5c)

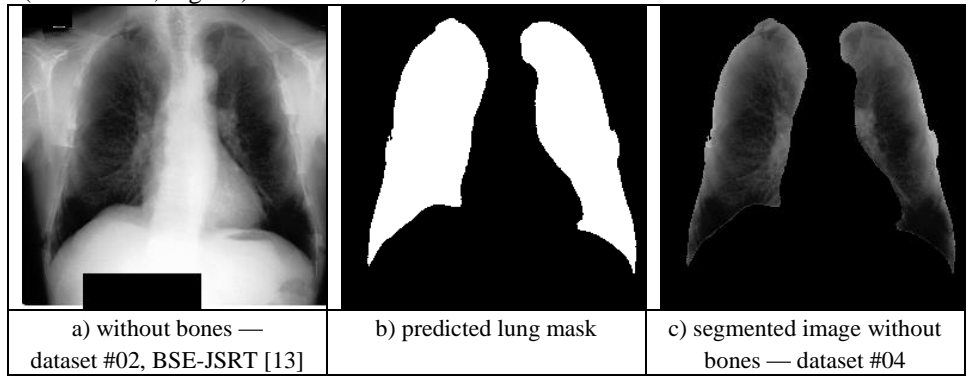

| a) without bones — dataset #02, BSE-JSRT [13] | b) predicted lung mask | c) segmented image without bones — dataset #04 |

**Fig. 5.** Example of the image without bones (inversed version of Fig.1b) (a), the correspondent lung mask predicted by machine learning (b), and its segmented version without bones (c).

### 3.5. Training and validation

Finally, the simple CNN was trained in GPU mode (NVIDIA Tesla K40c card) by means of Tensorflow machine learning framework [18] with regard to the 4 datasets: original JSRT dataset (dataset #01), original BSE-JSRT dataset, i.e. the same JSRT dataset, but without clavicle and rib shadows (dataset #02), JSRT dataset after segmentation (dataset #03), and BSE-JSRT dataset after segmentation (dataset #04).
The previous results (Fig. 6) clearly demonstrate the drastic difference in training and validation results between raw data, namely, the original JSRT dataset #01, and any of the pre-processed datasets #02, #03, or #04. Despite the original JSRT dataset #01

(red line on Fig. 6) does not show any sign of training at all (because of low image size and negligibly small nodule size), all of the pre-processed datasets #02, #03, or #04 (orange, dark blue, and blue lines on Fig. 6) have tendency to train and demonstrate high training accuracy and training low loss for the late stages in such simplified configuration even.

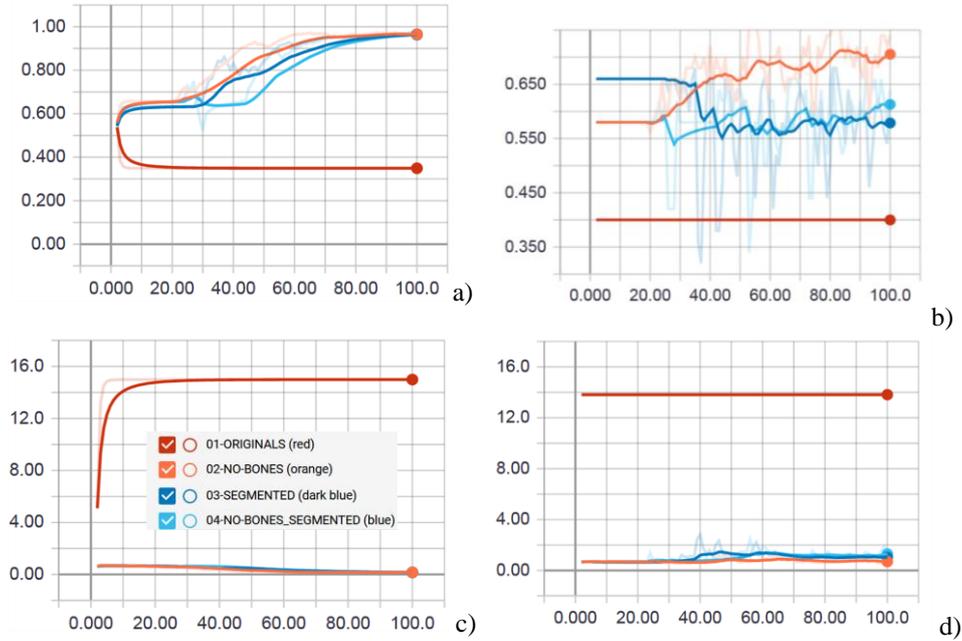

**Fig. 6.** Training accuracy (a), validation accuracy (b), training loss (c), and validation loss (d) for the original images, JSRT, dataset #01 (line 01), the images without bones, BSE-JSRT, dataset #02 (line 02), the segmented images, dataset #03 (line 03), and the segmented images without bones, dataset #04 (line 04).

The more careful analysis of training and validation curves of the pre-processed datasets #02, #03, or #04 (Fig. 7) allows to note that accuracy (and loss) obtained after training is much higher (lower) than ones observed after validation. This can be considered as the clear manifestation of overtraining with the overestimated accuracy and underestimated loss for the raw image data from the original JSRT dataset #01. Nevertheless, the pre-processed dataset without bones (dataset #02) demonstrates the much better accuracy and loss results in comparison to the other pre-processed datasets after lung segmentation (datasets #02 and #03).

## 5   Discussion and Conclusions

The results obtained demonstrate the usefulness of pre-processing techniques like bone shadow elimination and segmentation with the clear tendency to train in such simplified configuration even. The overtraining effect with the lower validation accuracy and higher validation loss in comparison to the training accuracy and loss is considered to be related with training to artifacts like the shape of the lungs and lung border pattern.

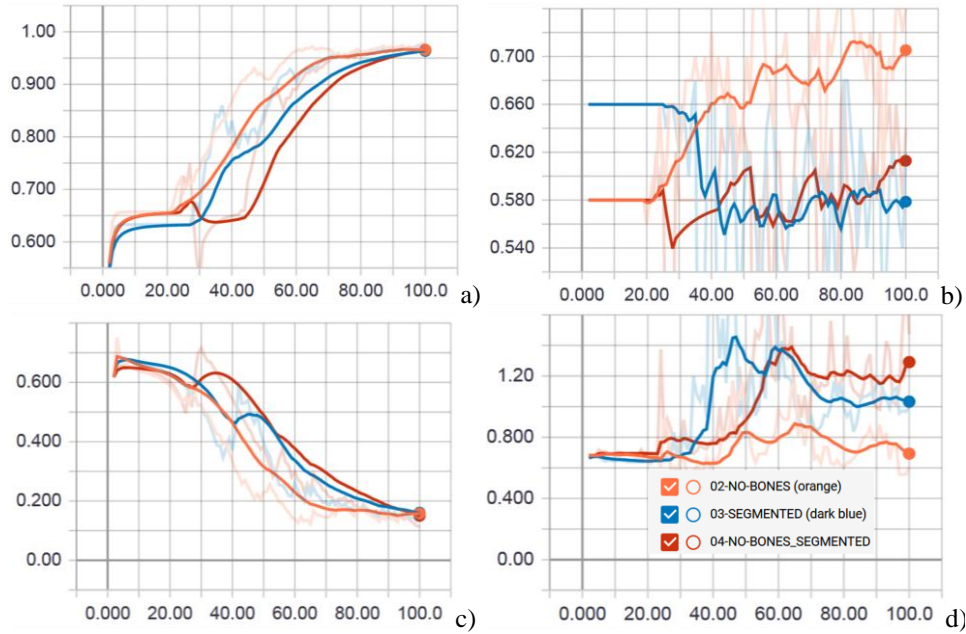

**Fig. 7.** Training accuracy (a), validation accuracy (b), training loss (c), and validation loss (d) for the images without bones (line 02), the segmented images (line 03), and the segmented images without bones (line 04).

The further potential for improvement consists in an increase of the investigated datasets:
- the image size from the current 256×256 values up to 1024×1024 (ChestXRay dataset) and 2048×2048 (JSRT dataset);
- the number of images from current 247 (JSRT and BSE-JSRT datasets) up to >1000 (MC dataset) and > 100 000 (ChestXRay dataset);
- by data augmentation with regard to lossy and lossless transformations [17].

The much bigger progress can be obtained from integration of many similar datasets from numerous hospitals around the world in the spirit of the open science data, volunteer data collection, data processing and computing [19].

In conclusion, the results demonstrate the high efficiency and usefulness of the considered pre-processing techniques in the simplified configuration even. It should be emphasized that the pre-processed dataset without bones (dataset #02) demonstrates the much better accuracy and loss results in comparison to the other pre-processed datasets after lung segmentation (datasets #02 and #03). That is why the additional reserve of development could be related with improvement of pre-processing algorithms for:
- lung segmentation on the basis of the bigger datasets with masks,

- bone shadow elimination on the basis of the more complicated semantic segmentation techniques applied not only to lungs and body parts outside of them (like heart, arms, etc.), but also to ribs and clavicles inside the lungs,
- training itself by means of increase of size and complexity of the deep learning network from the current miniature size of 7 layers up to >100 layers like in the current most accurate networks like CheXNet used for diagnostics of other diseases [5].

Training of the more complicated models on the bigger images are under work now and will be reported elsewhere [17]. In this connection the fine tuning of datasets and deep learning models should be taken into account, because it can play the crucial role for efficiency of the model used [20-22].

**Acknowledgments.** The work was partially supported by Huizhou Science and Technology Bureau and Huizhou University (Huizhou, P.R.China) in the framework of Platform Construction for China-Ukraine Hi-Tech Park Project # 2014C050012001.